\pdfoutput=1
\documentclass[11pt,a4paper]{article}
\usepackage[hyperref]{tacl2018} 
\usepackage{times,latexsym}
\usepackage{url}
\usepackage[T1]{fontenc}
\usepackage{amsmath,amsthm,amssymb}
\usepackage{mathrsfs}
\usepackage{threeparttable}
\usepackage{graphicx}
\usepackage{relsize}
\usepackage{booktabs}
\usepackage{multirow}
\usepackage{algorithm}
\usepackage[noend]{algpseudocode}
\usepackage{bm}
\usepackage{bbm}

\taclfinalcopy 

\newcommand{\secref}[1]{\S\ref{#1}}

\algnewcommand{\LeftComment}[1]{\State \(\triangleright\) #1}

\newcommand{\cf}{{\cal F}}
\newcommand{\cp}{{\cal P}}
\newcommand{\cq}{{\cal Q}}
\newcommand{\expe}{\mathop{{}\mathbb{E}}}

\newcommand{\adan}{\texttt{ADAN}}
\newcommand{\source}{\textsc{S{\smaller OURCE}}}
\newcommand{\target}{\textsc{T{\smaller ARGET}}}

\title{Adversarial Deep Averaging Networks\\for Cross-Lingual Sentiment Classification}

\author{Xilun Chen\textsuperscript{\normalfont\textdagger} \\ \texttt{xlchen@cs.cornell.edu}
\And
Yu Sun\textsuperscript{\normalfont\textdagger}\\ \texttt{ys646@cornell.edu}
\And
Ben Athiwaratkun\textsuperscript{\normalfont\textdaggerdbl}\\ \texttt{pa338@cornell.edu} \\
\AND
Claire Cardie\textsuperscript{\normalfont\textdagger} \\ \texttt{cardie@cs.cornell.edu}
\And
Kilian Weinberger\textsuperscript{\normalfont\textdagger} \\ \texttt{kqw4@cornell.edu}
\AND\vspace*{-5mm}\\
\textdagger Dept. of Computer Science, Cornell University, Ithaca NY, USA\\
\textdaggerdbl Dept. of Statistical Science, Cornell University, Ithaca NY, USA}

\date{}

\begin{document}

\maketitle

\begin{abstract}
  In recent years great success has been achieved in sentiment classification for English,
thanks in part to the availability of copious annotated resources.
Unfortunately, most languages do not enjoy such an abundance of labeled data.
To tackle the sentiment classification problem in low-resource languages without adequate annotated data, we propose an Adversarial Deep Averaging Network (\texttt{ADAN}\footnote{The source code of \adan{} is available at \url{https://github.com/ccsasuke/adan}}) to transfer the knowledge learned from labeled data on a resource-rich source language to low-resource languages where \textit{only unlabeled} data exists.
\texttt{ADAN} has two discriminative branches: a \textit{sentiment classifier} and an adversarial \textit{language discriminator}.
Both branches take input from a shared \textit{feature extractor} to learn hidden representations that are simultaneously indicative for the classification task and \textit{invariant} across languages. 
Experiments on Chinese and Arabic sentiment classification demonstrate that \texttt{ADAN} significantly outperforms state-of-the-art systems.
\end{abstract}

\section{Introduction}\label{sec:intro}
Many state-of-the-art models for sentiment classification~\cite{socher-EtAl:2013:EMNLP,iyyer-EtAl:2015:ACL-IJCNLP,tai-socher-manning:2015:ACL-IJCNLP} are supervised learning approaches that rely on the availability of an adequate amount of labeled training data.
For a few resource-rich languages including English, such labeled data is indeed available.
For the vast majority of languages, however, it is the norm that only a limited amount of annotated text exists.
Worse still, many low-resource languages have no labeled data at all.

To aid the creation of sentiment classification systems in such low-resource languages,
an active research direction is cross-lingual sentiment classification (CLSC) in which the abundant resources of a \emph{source} language (likely English, denoted as \source{}) are leveraged to produce sentiment classifiers for a \emph{target} language (\target{}).
In general, CLSC methods make use of general-purpose bilingual resources --- such as hand-crafted bilingual lexica or parallel corpora --- to alleviate or eliminate the need for task-specific \target{} annotations.
In particular, the bilingual resource of choice for the majority of previous CLSC models is a full-fledged Machine Translation (MT) system~\cite{wan2008using,wan2009co,lu2011joint,P16-1133}, a component that is expensive to obtain.
In this work, we propose a \textbf{language-adversarial training} approach that does not need a highly engineered MT system, and requires orders of magnitude less in terms of the size of parallel corpus.
Specifically, we propose an Adversarial Deep Averaging Network (\adan{}) that leverages a set of Bilingual Word Embeddings~\citep[BWEs,][]{zou-EtAl:2013:EMNLP} trained on bitexts, in order to eliminate the need for labeled \target{} training data\footnote{When not using any \target{} annotations, the setting is sometimes referred to as \emph{unsupervised} (in the target language) in the literature. Similarly, when some labeled data is used, it is called the \emph{semi-supervised} setting.}.

We introduce the \adan{} model in~\secref{sec:model}, and in~\secref{sec:experiments} evaluate \adan{} using English as the \source{} with two \target{} choices: Chinese and Arabic.
\adan{} is first compared to two baseline systems: 
i) one trained only on labeled \source{} data, relying on BWEs for cross-lingual generalization;
and ii) a domain adaptation method~\cite{ICML2012Chen_416} that views the two languages simply as two distinct domains.
We then validate \adan{} against two state-of-the-art CLSC methods:
iii) an approach that employs a powerful MT system,
and iv) the cross-lingual ``distillation'' approach of \newcite{P17-1130} that makes direct use of a parallel corpus (see \secref{sec:adanexp}).
In all cases, we find that \adan{} achieves statistically significantly better results.

We further investigate the semi-supervised setting, where a small amount of annotated \target{} data exists,
and show that \texttt{ADAN} continues to outperform the alternatives given the same amount of \target{} supervision (\secref{sec:semi-sup}).
We provide an analysis and visualization of \texttt{ADAN} (\secref{sec:visualization}), shedding light on how our approach manages to achieve its strong cross-lingual performance.
Additionally, we study the bilingual resource that \adan{} depends on, the Bilingual Word Embeddings, and demonstrate that \texttt{ADAN}'s performance is robust with respect to the choice of BWEs.
Furthermore, even without the pre-trained BWEs (i.e. using random initialized embeddings), \adan{} outperforms all but the state-of-the-art MT-based and distillation systems (\secref{sec:wordembexp}).
This makes \adan{} the first CLSC model that outperforms BWE-based baseline systems without relying on \emph{any} bilingual resources.

A final methodological contribution distinguishes \adan{} from previous adversarial networks for text classification~\cite{JMLR:v17:15-239}: 
\adan{} minimizes the Wasserstein distance~\cite{arjovsky2017wasserstein} between the feature distributions of \source{} and \target{} (\secref{sec:training}), which yields better performance and smoother training than the standard GRL training method~\citep[][see \secref{sec:hyperparameter}]{JMLR:v17:15-239}.

\section{The \adan{} Model}\label{sec:model}
\begin{figure}[!t]
  \centering
  \includegraphics[width=0.4\textwidth]{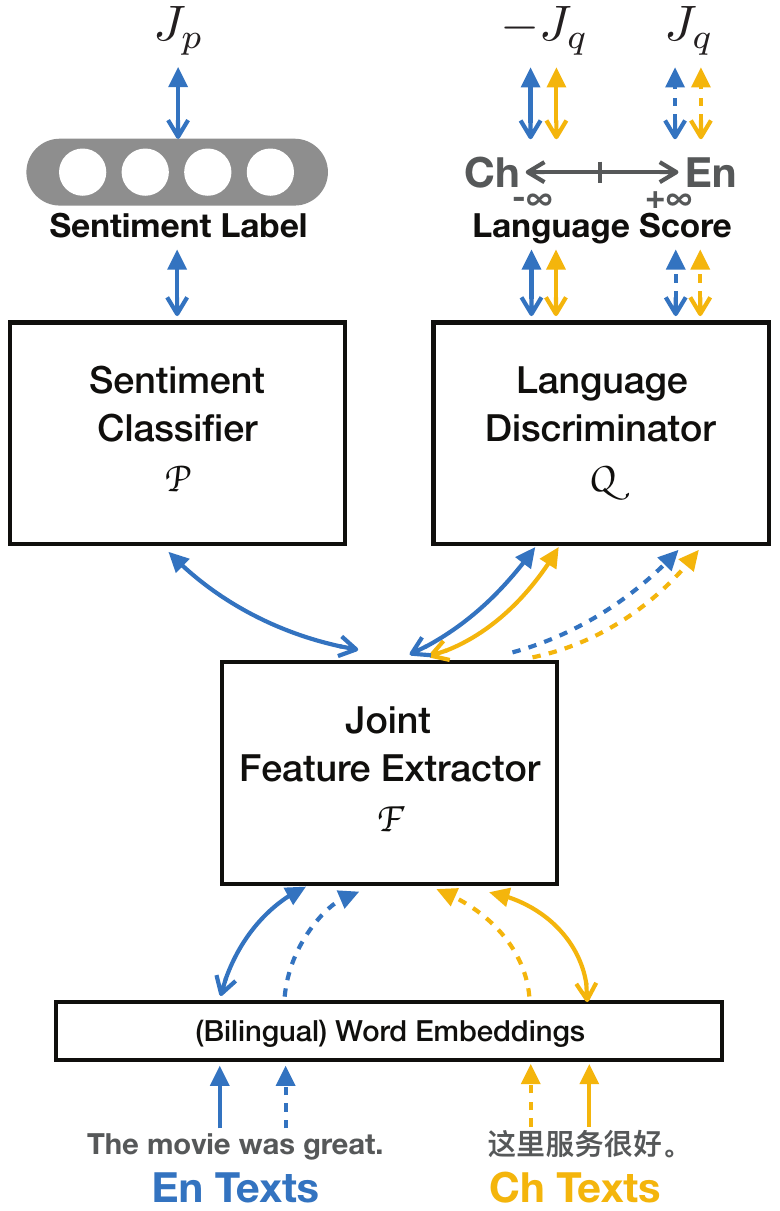}
  \caption{
  ADAN with Chinese as the target language.
  The lines illustrate the training flows and the arrows indicate forward and/or backward passes.
  Blue lines show the flow for English samples while Yellow ones are for Chinese.
  $J_p$ and $J_q$ are the training objectives of $\cp$ and $\cq$, respectively (\secref{sec:training}).
  The parameters of $\cf$, $\cp$ and the embeddings are updated together (solid lines).
  The parameters of $\cq$ are updated using a separate optimizer (dotted lines) due to its adversarial objective.
  }
  \label{fig:modeloverview}
\end{figure}

The central hypothesis of \adan{} is that an ideal model for CLSC should learn features that both perform well on sentiment classification for the \source{}, and are invariant with respect to the shift in language.
Therefore, as shown in Figure~\ref{fig:modeloverview}, \adan{} has a joint \emph{feature extractor} $\cf$ which aims to learn features that aid prediction of the \emph{sentiment classifier} $\cp$, and \textbf{hamper} the \emph{language discriminator} $\cq$, whose goal is to identify whether an input text is from \source{} or \target{}.
The intuition is that if a well-trained $\cq$ cannot tell the language of a given input using the features extracted by $\cf$, those features are effectively language-invariant.
$\cq$ is hence \emph{adversarial} since it does its best to identify language from learned features, yet good performance from $\cq$ indicates that \adan{} is not successful in learning language-invariant features.
Upon successful \adan{} training, $\cf$ should have learned features discriminative for sentiment classification, and at the same time providing no information for the adversarial $\cq$ to guess the language of a given input.

As seen in Figure~\ref{fig:modeloverview}, \adan{} is exposed to both \source{} and \target{} texts during training.
Unlabeled \source{} (blue lines) and \target{} (yellow lines) data go through the language discriminator, while only the labeled \source{} data pass through the sentiment classifier\footnote{``Unlabeled'' and ``labeled'' refer to sentiment labels; all texts are assumed to have the correct language label.}. 
The feature extractor and the sentiment classifier are then used for \target{} texts at test time.
In this manner, we can train \texttt{ADAN} with labeled \source{} data and only unlabeled \target{} text.
When some labeled \target{} data exist, \adan{} could naturally be extended to take advantage of that for improved performance (\secref{sec:semi-sup}).

\subsection{Network Architecture}\label{sec:modelarch}
As illustrated in Figure~\ref{fig:modeloverview},
\texttt{ADAN} is a feed-forward network with two branches.
There are three main components in the network,
a joint \textit{feature extractor} $\mathcal{F}$ that maps an input sequence $x$ to the shared feature space,
a \textit{sentiment classifier} $\mathcal{P}$ that predicts the label for $x$ given the feature representation $\mathcal{F}(x)$,
and a \textit{language discriminator} $\mathcal{Q}$ that also takes $\cf(x)$ but predicts a scalar score indicating whether $x$ is from \source{} or \target{}.

An input document is modeled as a sequence of words $x=w_1,\dots,w_n$,
where each $w$ is represented by its word embedding $v_w$~\cite{P10-1040}.
For improved performance, pre-trained bilingual word embeddings \citep[BWEs,][]{zou-EtAl:2013:EMNLP,gouws2015bilbowa} can be employed to induce bilingual distributed word representations so that similar words are closer in the embedded space regardless of language.

A parallel corpus is often required to train high-quality BWEs,
making \texttt{ADAN} implicitly dependent on the bilingual corpus.
However, compared to the MT systems used in other CLSC methods, training BWEs only requires one to two orders of magnitude less parallel data, and some methods only take minutes to train on a consumer CPU~\cite{gouws2015bilbowa}, while state-of-the-art MT systems need days to weeks for training on multiple GPUs.
Moreover, even with randomly initialized embeddings, \adan{} can still outperform some baseline methods that use pre-trained BWEs (\secref{sec:wordembexp}).
Another possibility is to take advantage of the recent work that trains BWEs with no bilingual supervision~\cite{lample2018word}.

We adopt the Deep Averaging Network (\texttt{DAN}) by \newcite{iyyer-EtAl:2015:ACL-IJCNLP} for the feature extractor $\cf$.
We choose \texttt{DAN} for its simplicity to illustrate the effectiveness of our language-adversarial training framework, but other architectures can also be used for the feature extractor (\secref{sec:f_arc_exp}).
For each document, \texttt{DAN} takes the arithmetic mean of the word vectors as input, and passes it through several fully-connected layers until a softmax for classification.
In \texttt{ADAN}, $\cf$ first calculates the average of the word vectors in the input sequence,
then passes the average through a feed-forward network with ReLU nonlinearities.
The activations of the last layer in $\cf$ are considered the extracted features for the input and are then passed on to $\cp$ and $\cq$.
The sentiment classifier $\cp$ and the language discriminator $\cq$ are standard feed-forward networks. 
$\cp$ has a softmax layer on top for sentiment classification and $\cq$ ends with a linear layer of output width 1 to assign a language identification score\footnote{$\cq$ simply tries to maximize scores for \source{} texts and minimize for \target{}, and the scores are not bounded.}.

\subsection{Adversarial Training}\label{sec:training}
For clarity, we first introduce an \adan{} variant where training is done using Gradient Reversal Layer~\cite{JMLR:v17:15-239}, which is denoted as \adan{}-GRL.
\adan{}-GRL employs standard adversarial training techniques in previous literature~\cite{JMLR:v17:15-239}, but as we will detail later in this section, the training of \adan{}-GRL is less stable and the performance is worse compared to our \adan{} model (see \secref{sec:hyperparameter} for empirical results).

Specifically, in \adan{}-GRL, $\cq$ is a binary classifier with a sigmoid layer on top so that the language identification score is always between 0 and 1 and is interpreted as the probability of whether an input text $x$ is from \source{} or \target{} given its hidden features $\cf(x)$.
For training, $\cq$ is connected to $\cf$ via a Gradient Reversal Layer~\cite{icml2015_ganin15}, which preserves the input during the a forward pass but multiplies the gradients by $-\lambda$ during a backward pass.
$\lambda$ is a hyperparameter that balances the effects that $\cp$ and $\cq$ have on $\cf$ respectively.
This way, the entire network can be trained in its entirety using standard backpropagation.

Unfortunately, researchers have found that the training of $\cf$ and $\cq$ in \adan{}-GRL might not be fully in sync~\cite{icml2015_ganin15}, and efforts need to be made to coordinate the adversarial training.
This is achieved by setting $\lambda$ to a non-zero value only once out of $k$ batches as in practice we observe that $\cf$ trains faster than $\cq$.
Here, $k$ is another hyperparameter that coordinates the training of $\cf$ and $\cq$.
When $\lambda=0$, the gradients from $\cq$ will not be back-propagated to $\cf$. This allows $\cq$ more iterations to adapt to $\cf$ before $\cf$ makes another adversarial update.

To illustrate the limitations of \adan{}-GRL and motivate the formal introduction of our \adan{} model, consider the distribution of the joint hidden features $\cf$ for both \source{} and \target{} instances:
\begin{align*}
    P_\cf^{src} &\triangleq P(\cf(x) | x\in \source{}) \\
    P_\cf^{tgt} &\triangleq P(\cf(x) | x\in \target{}) 
\end{align*}

In order to learn language-invariant features, \adan{} trains $\cf$ to make these two distributions as close as possible for better cross-lingual generalization.
In particular, as argued by~\newcite{arjovsky2017wasserstein}, previous approaches to 
training adversarial networks such as \adan{}-GRL are equivalent to minimizing the Jensen-Shannon divergence between two distributions, in our case $P_\cf^{src}$ and  $P_\cf^{tgt}$.
And because the Jensen-Shannon divergence suffers from discontinuities, providing less useful gradients for training $\cf$, \newcite{arjovsky2017wasserstein} propose instead to minimize the Wasserstein distance and demonstrate its improved stability for hyperparameter selection.

\begin{algorithm}[t]
\small
\begin{algorithmic}[1]
\Require
labeled \source{} corpus $\mathbb{X}_{src}$; unlabeled \target{} corpus $\mathbb{X}_{tgt}$; Hyperpamameter $\lambda > 0$, $k \in \mathbb{N}$, $c > 0$.
\Repeat
\LeftComment{$\cq$ iterations}
\For{$qiter = 1$ to $k$}
\State Sample unlabeled batch $\bm{x}_{src} \sim \mathbb{X}_{src}$
\State Sample unlabeled batch $\bm{x}_{tgt} \sim \mathbb{X}_{tgt}$
\State $\bm{f}_{src} = \cf(\bm{x}_{src})$
\State $\bm{f}_{tgt} = \cf(\bm{x}_{tgt})$ \Comment{feature vectors}
\State $loss_q = - \cq(\bm{f}_{src}) + \cq(\bm{f}_{tgt})$ \Comment{Eqn~(\ref{eqn:jq})}
\State Update $\cq$ parameters to minimize $loss_q$
\State $ClipWeights(\cq, -c, c)$
\EndFor

\LeftComment{Main iteration}
\State Sample labeled batch $(\bm{x}_{src},\bm{y}_{src})  \sim \mathbb{X}_{src}$
\State Sample unlabeled batch $\bm{x}_{tgt} \sim \mathbb{X}_{tgt}$
\State $\bm{f}_{src} = \cf(\bm{x}_{src})$
\State $\bm{f}_{tgt} = \cf(\bm{x}_{tgt})$
\State $loss = L_p(\cp(\bm{f}_{src}); \bm{y}_{src}) + \lambda (\cq(\bm{f}_{src}) - \cq(\bm{f}_{tgt}))$ \Comment{Eqn~(\ref{eqn:jf})}

\State Update $\cf$, $\cp$ parameters to minimize $loss$
\Until{convergence}

\end{algorithmic}
\caption{\adan{} Training}
\label{alg:training}
\end{algorithm}

As a result, departing from the previous \adan{}-GRL training method, in our \adan{} model, we minimize the Wasserstein distance $W$ between $P_\cf^{src}$ and $P_\cf^{tgt}$ according to the Kantorovich-Rubinstein duality~\cite{villani2008optimal}:
\begin{align}
    &W(P_\cf^{src}, P_\cf^{tgt}) = \label{eqn:wasserstein}\\
    &\sup_{\lVert g \rVert_L \leq 1} \expe_{f(x)\sim P_\cf^{src}} \left[ g(f(x)) \right] - \expe_{f(x')\sim P_\cf^{tgt}} \left[ g(f(x')) \right] \nonumber
\end{align}
where the supremum (maximum) is taken over the set of all 1-Lipschitz\footnote{A function $g$ is 1-Lipschitz iff $|g(x)-g(y)|\leq |x-y|$ for all $x$ and $y$.} functions $g$.
In order to (approximately) calculate $W(P_\cf^{src},P_\cf^{tgt})$,
we use the language discriminator $\cq$ as the function $g$ in (\ref{eqn:wasserstein}),
whose objective is then to seek the supremum in (\ref{eqn:wasserstein}).
To make $\cq$ a Lipschitz function (up to a constant), the parameters of $\cq$ are always clipped to a fixed range $[-c, c]$.
Let $\cq$ be parameterized by $\theta_q$, then the objective $J_q$ of $\cq$ becomes:
\begin{align}
    &J_q(\theta_f) \equiv \label{eqn:jq}\\
    &\max_{\theta_q} \expe_{\cf(x)\sim P_\cf^{src}} \left[ \cq(\cf(x)) \right] - \expe_{\cf(x')\sim P_\cf^{tgt}} \left[ \cq(\cf(x')) \right] \nonumber
\end{align}
Intuitively, $\cq$ tries to output higher scores for \source{} instances and lower scores for \target{}.
More formally, $J_q$ is an approximation of the Wasserstein distance between $P_\cf^{src}$ and $P_\cf^{tgt}$ in (\ref{eqn:wasserstein}).

For the sentiment classifier $\cp$ parameterized by $\theta_p$,
we use the traditional cross-entropy loss, denoted as $L_p(\hat{y},y)$,
where $\hat{y}$ and $y$ are the predicted label distribution and the true label, respectively.
$L_p$ is the negative log-likelihood that $\cp$ predicts the correct label.
We therefore seek the minimum of the following loss function for $\cp$:
\begin{equation}
  J_p(\theta_f)\equiv \min_{\theta_p}\expe_{(x,y)} \left[ L_p(\cp(\cf(x)),y)\right]
  \label{eqn:jp}
\end{equation}

Finally, the joint feature extractor $\cf$ parameterized by $\theta_f$ strives to minimize both the sentiment classifier loss $J_p$ and $W(P_\cf^{src}, P_\cf^{tgt}) = J_q$:
\begin{align}
  J_f \equiv \min_{\theta_f} J_p(\theta_f) + \lambda J_q(\theta_f) \label{eqn:jf}
\end{align}
where $\lambda$ is a hyper-parameter that balances the two branches $\cp$ and $\cq$.

As proved by~\newcite{arjovsky2017wasserstein} and observed in our experiments (\secref{sec:hyperparameter}),
minimizing the Wasserstein distance is much more stable w.r.t.\ hyperparameter selection compared to \adan{}-GRL, saving the hassle of carefully varying $\lambda$ during training~\cite{icml2015_ganin15}.
In addition, \adan{}-GRL needs to laboriously coordinate the alternating training of the two competing components by setting the hyperparameter $k$, which indicates the number of iterations one component is trained before training the other.
The performance can degrade substantially if $k$ is not properly set.
In our case, however, delicate tuning of $k$ is no longer necessary since $W(P_\cf^{src}, P_\cf^{tgt})$ is approximated by maximizing (\ref{eqn:jq});
thus, training $\cq$ to optimum using a large $k$ can provide better performance (but is slower to train).
In our experiments,
we fix $\lambda=0.1$ and $k=5$ for all experiments (train 5 $\cq$ iterations per $\cf$ and $\cp$ iteration), and the performance is stable over a large set of hyperparameters (\secref{sec:hyperparameter}).

\adan{} training is depicted in Algorithm~\ref{alg:training}.

\begin{table*}[ht]
  \centering
  \small
  \begin{threeparttable}
    \begin{tabular}{ l  l  c  c }
      \toprule
      \multirow{2}{*}{Methodology}	& \multirow{2}{*}{Approach}	& \multicolumn{2}{c}{Accuracy}\\
      \cmidrule{3-4}
      &&   Chinese  & Arabic\\
      \midrule
      \multirow{2}{*}{Train-on-\source-only}	& Logistic Regression &   30.58\%	& 45.83\%\\
      & \texttt{DAN}      &   29.11\% & 48.00\% \\ 
      \midrule
      Domain Adaptation	& mSDA~\cite{ICML2012Chen_416}              &   31.44\% & 48.33\%\\
      \midrule
      \multirow{2}{*}{Machine Translation}	& Logistic Regression + MT      &   34.01\% & 51.67\%\\
      & \texttt{DAN} + MT   &   39.66\% & 52.50\% \\ 
      \midrule
      \multirow{2}{*}{CLD-based CLTC}  & CLD-KCNN~\cite{P17-1130} & 40.96\% & 52.67\%\tnote{\textdagger} \\
      & CLDFA-KCNN~\cite{P17-1130} & 41.82\% & 53.83\%\tnote{\textdagger} \\
      \midrule
      Ours & \texttt{ADAN} 
      &   \textbf{42.49\%}\smaller{$\pm 0.19\%$} & \textbf{54.54\%}\smaller{$\pm 0.34\%$}\\
      \bottomrule
    \end{tabular}
    \begin{tablenotes}
    \item[\textdagger] As \newcite{P17-1130} did not report results for Arabic, these numbers are obtained based on our reproduction using their code.
    \end{tablenotes}
  \end{threeparttable}
  \caption{\texttt{ADAN} performance for Chinese (5-cls) and Arabic (3-cls) sentiment classification without using labeled \target{} data.
  All systems but the CLD ones use BWE to map \source{} and \target{} words into the same space. 
  CLD-based CLTC represents cross-lingual text classification methods based on cross-lingual distillation~\cite{P17-1130} and is explained in \secref{sec:adanexp}.
  For \adan{}, average accuracy and standard errors over five runs are shown.
  Bold numbers indicate statistical significance over all baseline systems with $p<0.05$ under a One-Sample T-Test.
  As a comparison, the supervised \emph{English} accuracy of our \adan{} model is $58.7\%$ (5-class) and $75.6\%$ (3-class).
}
\label{tab:adanexp}
\end{table*}

\section{Experiments and Discussions}\label{sec:experiments}

To demonstrate the effectiveness of our model, we experiment on Chinese and Arabic sentiment classification, using English as \source{} for both.
For all data used in experiments,
tokenization is done using Stanford CoreNLP~\cite{manning-EtAl:2014:P14-5}.

\subsection{Data}

\noindent
\textbf{Labeled English Data.} We use a balanced dataset of $700k$ Yelp reviews from~\newcite{zhang2015character} with their ratings as labels (scale 1-5).
We also adopt their train-validation split: $650k$ reviews for training and $50k$ form a validation set.

\noindent 
\textbf{Labeled Chinese Data.} Since \texttt{ADAN} does not require labeled Chinese data for training, this annotated data is solely used to validate the performance of our model.
$10k$ balanced Chinese hotel reviews from~\newcite{Lin-Lei-Wu-Li:2015:PACLIC} are used as validation set for model selection and parameter tuning.
The results are reported on a separate test set of another $10k$ hotel reviews.
For Chinese, the data are annotated with $5$ labels (1-5).

\noindent
\textbf{Unlabeled Chinese Data.} For the unlabeled \target{} data used in training \texttt{ADAN}, we use another $150k$ unlabeled Chinese hotel reviews.

\noindent\textbf{English-Chinese Bilingual Word Embeddings.}
For Chinese,
we used the pre-trained bilingual word embeddings (BWE) by~\newcite{zou-EtAl:2013:EMNLP}. Their work provides 50-dimensional embeddings for $100k$ English words and another set of $100k$ Chinese words.
See \secref{sec:wordembexp} for more experiments and discussions.

\noindent\textbf{Labeled Arabic Data.}
We use the BBN Arabic Sentiment Analysis dataset~\cite{mohammadSK2015} for Arabic sentiment classification.
The dataset contains 1200 sentences (600 validation + 600 test) from social media posts annotated with 3 labels ($-$, $0$, $+$).
The dataset also provides machine translated text to English.
Since the label set does not match with the English dataset,  we map all the rating 4 and 5 English instances to $+$ and the rating 1 and 2 instances to $-$,
while the rating 3 sentences are converted to $0$.

\noindent\textbf{Unlabeled Arabic Data.}
For Arabic, no additional unlabeled data is used.
We only use the text from the validation set (without labels) during training.

\noindent\textbf{English-Arabic Bilingual Word Embeddings.}
For Arabic, we train a 300d BilBOWA BWE~\cite{gouws2015bilbowa} on the United Nations corpus~\cite{ziemski2016united}.

\subsection{Cross-Lingual Sentiment Classification}\label{sec:adanexp}
Our main results are shown in Table~\ref{tab:adanexp},
which shows very similar trends for Chinese and Arabic.
Before delving into discussions on the performance of \adan{} compared to various baseline systems in the following paragraphs, we begin by clarifying the bilingual resources used in all the methods.
Note first that in all of our experiments, traditional features like bag of words cannot be directly used since \source{} and \target{} have completely different vocabularies.
Therefore, unless otherwise specified, BWEs are used as the input representation for all systems to map words from both \source{} and \target{} into the same feature space.
(The only exceptions are the CLD-based CLTC systems of \newcite{P17-1130} explained later in this section, which directly make use of a parallel corpus instead of relying on BWEs.)
The same BWEs are adopted in all systems that utilize BWEs.

\paragraph{Train-on-\source{}-only baselines}
We start by considering two baselines that train only on the \source{} language, English, and rely solely on the BWEs to classify the \target{}.
The first variation uses a standard supervised learning algorithm, Logistic Regression (LR), shown in Row 1 in Table~\ref{tab:adanexp}.
In addition, we evaluate a non-adversarial variation of \texttt{ADAN}, just the \texttt{DAN} portion of our model (Row 2),
which is one of the modern neural models for sentiment classification.
We can see from Table~\ref{tab:adanexp} that,
in comparison to \texttt{ADAN} (bottom line), the train-on-\source{}-only baselines perform poorly.
This indicates that BWEs by themselves do not suffice to transfer knowledge of English sentiment classification to \target{}.

\paragraph{Domain Adaptation baselines}
We next compare \texttt{ADAN} with domain adaptation baselines, since domain adaptation can be viewed as a generalization of the cross-lingual task.
Nonetheless, the divergence between languages is much more significant than the divergence between two domains, which are typically two product categories in practice.
Among domain adaptation methods, the widely-used TCA~\cite{pan2011domain} did not work since it required quadratic space in terms of the number of samples (650k).
We thus compare to mSDA~\cite{ICML2012Chen_416}, a very effective method for cross-domain sentiment classification on Amazon reviews.
However, as shown in Table~\ref{tab:adanexp} (Row 3), mSDA did not perform competitively.
We speculate that this is because many domain adaptation models including mSDA were designed for the use of bag-of-words features,
which are ill-suited in our task where the two languages have completely different vocabularies.
In summary, this suggests that even strong domain adaptation algorithms cannot be used out of the box with BWEs for the CLSC task. 

\paragraph{Machine Translation baselines}
We then evaluate \texttt{ADAN} against Machine Translation baselines (Rows 4-5) that
(1) translate the \target{} text into English and then
(2) use the better of the train-on-\source{}-only models for sentiment classification.
Previous studies~\cite{banea2008multilingual,salameh-mohammad-kiritchenko:2015:NAACL-HLT} on sentiment classification for Arabic and European languages claim this MT approach to be very competitive and find that it can sometimes match the state-of-the-art system trained on that language.
For Chinese, where translated text was not provided, we use the commercial Google Translate engine\footnote{\url{https://translate.google.com}},
which is highly engineered, trained on enormous resources, and arguably one of the best MT systems currently available.
As shown in Table~\ref{tab:adanexp}, our \texttt{ADAN} model substantially outperforms the MT baseline on both languages,
indicating that our adversarial model can successfully perform cross-lingual sentiment classification without any annotated data in the target language. 

\paragraph{Cross-lingual Text Classification baselines}
Finally, we conclude ADAN's effectiveness by comparing against a state-of-the-art cross-lingual text classification (CLTC) method~\cite{P17-1130}, as sentiment classification is one type of text classification.
They propose a cross-lingual distillation (CLD) method that makes use of soft \source{} predictions on a parallel corpus to train a \target{} model (CLD-KCNN).
They further propose an improved variant (CLDFA-KCNN) that utilizes adversarial training to bridge the \emph{domain} gap between the labeled and unlabeled texts within the source and the target language, similar to the adversarial domain adaptation by~\newcite{JMLR:v17:15-239}.
In other words, CLDFA-KCNN consists of three conceptual adaptation steps:
(i) Domain adaptation from source-language labeled texts to source-language unlabeled texts using adversarial training;
(ii) Cross-lingual adaptation using distillation;
and (iii) Domain adaptation in the target language from unlabeled texts to the test set.
Note, however, \newcite{P17-1130} use adversarial training for domain adaptation within a single language vs.\ our work that uses adversarial training directly for cross-lingual generalization.

As shown in Table~\ref{tab:adanexp}, ADAN significantly outperforms both variants of CLD-KCNN and achieves a new state of the art performance, indicating that our direct use of adversarial neural nets for cross-lingual adaptation can be more effective than chaining three adaptation steps as in CLDFA-KCNN.
This is the case in spite of the fact that \adan{} does not explicitly separate language variation from domain variation.
In fact, the monolingual data we use for the source and target languages is indeed from different domains. 
\adan{}'s performance suggests that it could potentially bridge the divergence introduced by both sources of variation in one shot.

\paragraph{Supervised \source{} accuracy}
By way of comparison, it is also instructive to compare \adan{}'s ``transferred'' accuracy on the \target{} with its (supervised) performance on the \source{}.   
As shown in the caption of Table~\ref{tab:adanexp}, \adan{} achieves 58.7\% accuracy on English for the 5-class English-Chinese setting, and 75.6\% for the 3-class English-Arabic setting.
The \source{} accuracy for the \texttt{DAN} baselines (Rows 2 and 5) is similar to the \source{} accuracy of \adan{}.

\subsection{Analysis and Discussion}

Since the Arabic dataset is small,
we choose Chinese as an example for our further analysis.

\subsubsection{Semi-supervised Learning}\label{sec:semi-sup}
\begin{figure}[tb]
  \centering
  \includegraphics[width=0.48\textwidth]{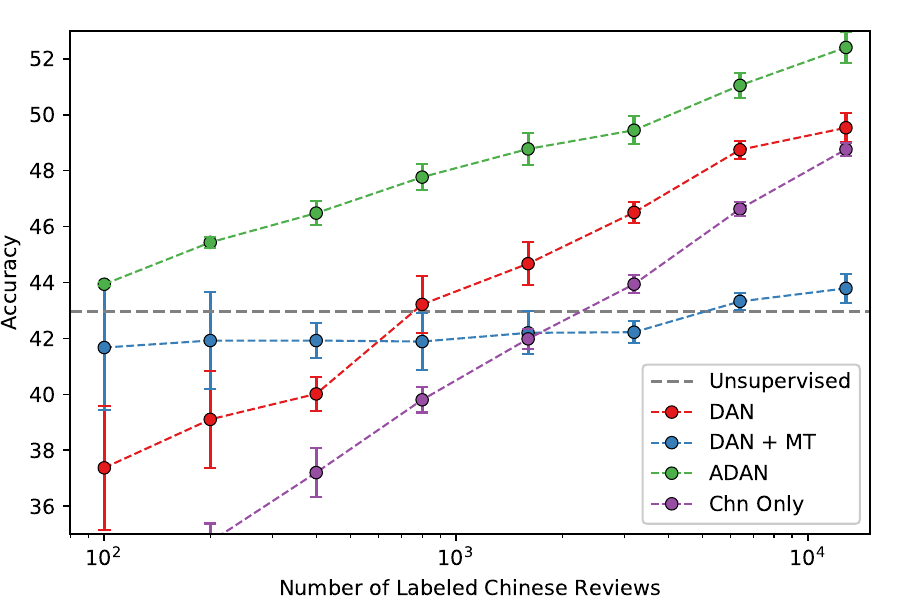}
  \caption{\texttt{ADAN} performance and standard deviation for Chinese in the semi-supervised setting when using various amount of labeled Chinese data.}
  \label{fig:semi-sup}
\end{figure}

\begin{figure*}[t]
  \centering
  \includegraphics[width=\textwidth]{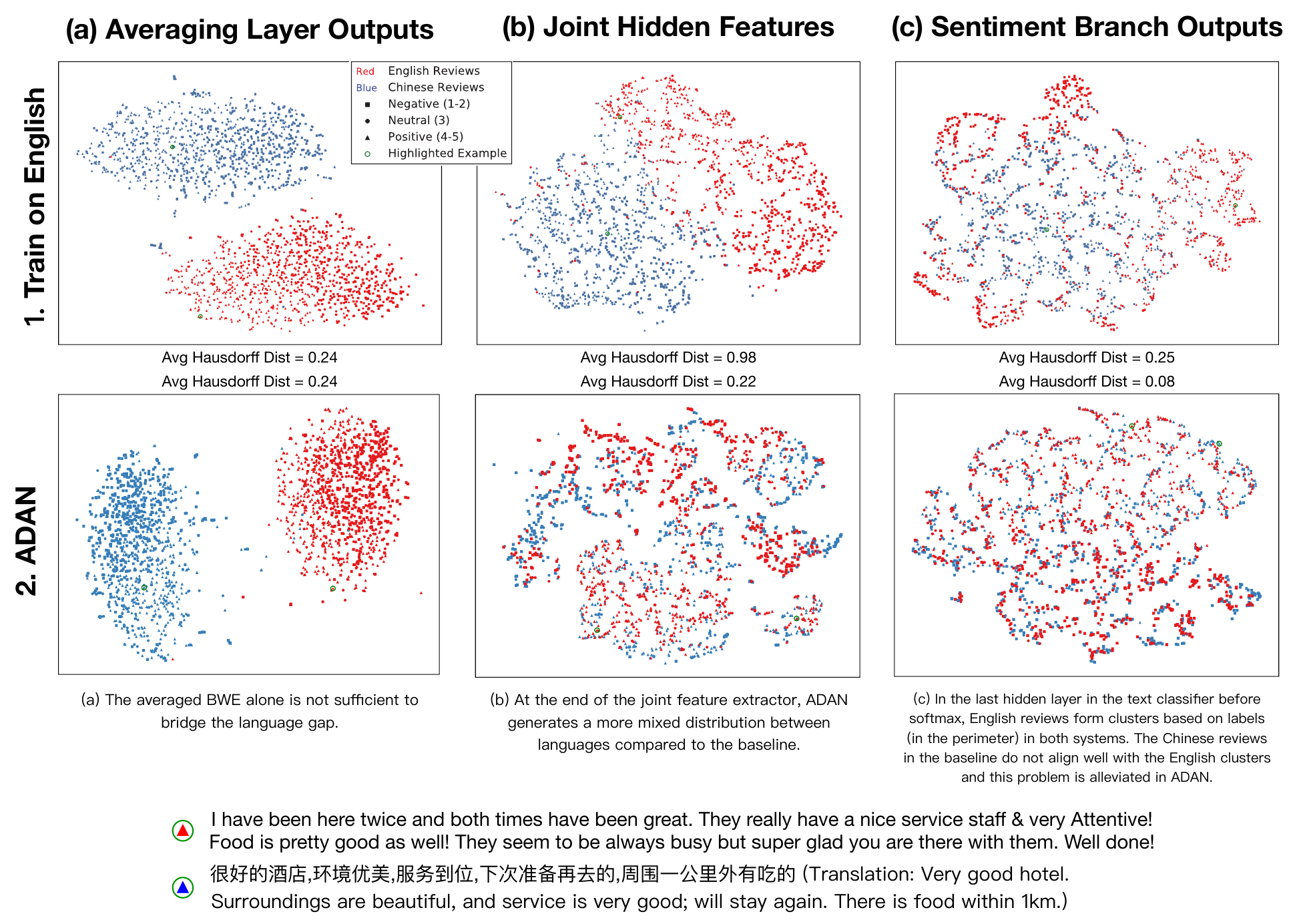}
  \caption{
    t-SNE visualizations of activations at various layers for the train-on-\source{}-only baseline model (top) and \texttt{ADAN} (bottom).
    The distributions of the two languages are brought much closer in \texttt{ADAN} as they are represented deeper in the network (left to right) measured by the Averaged Hausdorff Distance (see text).
    The green circles are two 5-star example reviews (shown below the figure) that illustrate how the distribution evolves (zoom in for details).
  }
  \label{fig:tsne}
\end{figure*}

In practice, it is usually not very difficult to obtain at least a small amount of annotated data.
\texttt{ADAN} can be readily adapted to exploit such extra labeled data in the target language,
by letting those labeled instances pass through the sentiment classifier $\cp$ as the English samples do during training.
We simulate this semi-supervised scenario by adding labeled Chinese reviews for training.
We start by adding 100 labeled reviews and keep doubling the number until 12800.
As shown in Figure~\ref{fig:semi-sup},
when adding the same number of labeled reviews, \texttt{ADAN} can better utilize the extra supervision and outperform the \texttt{DAN} baseline trained with combined data,
as well as the supervised \texttt{DAN} using only labeled Chinese reviews.
The margin is naturally decreasing as more supervision is incorporated, but \texttt{ADAN} is still superior when adding 12800 labeled reviews.
On the other hand, the \texttt{DAN} with translation baseline seems unable to effectively utilize the added supervision in Chinese, and the performance only starts to show a slightly increasing trend when adding 6400 or more labeled reviews.
One possible reason is that when adding to the training data a small number of English reviews translated from the labeled Chinese data,
the training signals they produce might be lost in the vast number of English training samples,
and thus not effective in improving performance.
Another potentially interesting find is that it seems a very small amount of supervision (e.g. 100 labels) could significantly help \texttt{DAN}.
However, with the same number of labeled reviews, \texttt{ADAN} still outperforms the \texttt{DAN} baseline.

\subsubsection{Qualitative Analysis and Visualizations}\label{sec:visualization}
To qualitatively demonstrate how \texttt{ADAN} bridges the distributional discrepancies between English and Chinese instances,
t-SNE~\cite{van2008visualizing} visualizations of the activations at various layers are shown in Figure~\ref{fig:tsne}.
We randomly select 1000 reviews from the Chinese and English validation sets respectively,
and plot the t-SNE of the hidden node activations at three locations in our model: the averaging layer, the end of the joint feature extractor, and the last hidden layer in the sentiment classifier just prior to softmax.
The train-on-English model is the \texttt{DAN} baseline in Table~\ref{tab:adanexp}.
Note that there is actually only one ``branch'' in this baseline model,
but in order to compare to \texttt{ADAN},
we conceptually treat the first three layers as the feature extractor.

Figure~\ref{fig:tsne}a shows that BWEs alone do not suffice to bridge the gap between the distributions of the two languages.
To shed more light on the surprisingly clear separation given that individual words have a mixed distribution in both languages (not shown in figure), we first try to isolate the content divergence from the language divergence.
In particular, the English and Chinese reviews are not translations of each other, and in fact may even come from different domains.
Therefore, the separation could potentially come from two sources: the content divergence between the English and Chinese reviews, and the language divergence of how words are used in the two languages.
To control for content divergence, we tried plotting (not shown in figure) the average word embeddings of 1000 random Chinese reviews and their machine translations into English using t-SNE, and surprisingly the clear separation was still present.
There are a few relatively short reviews that reside close to their translations, but the majority still form two language islands.
(The same trend persists when we switch to a different set of pre-trained BWEs, and when we plot a similar graph for English-Arabic.)
When we remove stop words (the most frequent word types in both languages), the two islands finally start to become slightly closer with less clean boundaries, but the separation remains clear.
We think this phenomenon is interesting, and a thorough investigation is out of the scope of this work.
We hypothesize that at least in certain distant language pairs such as English-Chinese\footnote{In a personal correspondence with Ahmed Elgohary, he did not observe the same phenomenon between English and French.}, the divergence between languages may not only be determined by word semantics, but also largely depends on how words are used.

Furthermore, we can see in Figure~\ref{fig:tsne}b that the distributional discrepancies between Chinese and English are significantly reduced after passing through the joint feature extractor ($\cf$).
The learned features in \texttt{ADAN} bring the distributions in the two languages dramatically closer compared to the monolingually trained baseline.
This is shown via the Averaged Hausdorff Distance~\citep[AHD,][]{shapiro2004hausdorff},
which measures the distance between two sets of points.
The AHD between the English and Chinese reviews is provided for all sub-plots in Figure~\ref{fig:tsne}.

Finally, when looking at the last hidden layer activations in the sentiment classifier of the baseline model (Figure~\ref{fig:tsne}c),
there are several notable clusters of red dots (English data) that roughly correspond to the class labels.
These English clusters are the areas where the classifier is the most confident in making decisions.
However, most Chinese samples are not close to one of those clusters due to the distributional divergence and may thus cause degraded classification performance in Chinese.
On the other hand, the Chinese samples are more in line with the English ones in \adan{},
which results in the accuracy boost over the baseline model.
In Figure~\ref{fig:tsne}, a pair of similar English and Chinese 5-star reviews is highlighted to visualize how the distribution evolves at various points of the network.
We can see in \ref{fig:tsne}c that the highlighted Chinese review gets close to the ``positive English cluster'' in \texttt{ADAN},
while in the baseline, it stays away from dense English clusters where the sentiment classifier trained on English data is not confident to make predictions.

\subsubsection{Impact of Bilingual Word Embeddings}\label{sec:wordembexp}
\begin{table}[bt]
  \centering
  \begin{tabular}{lccc}
    \toprule
    Model	&	Random	 &	BilBOWA	& \citeauthor{zou-EtAl:2013:EMNLP} \\
    \midrule
    \texttt{DAN}	& 21.66\%	& 28.75\%	& 29.11\%\\
    \texttt{DAN}+MT	& 37.78\%	& 38.17\%	& 39.66\%\\
    \texttt{ADAN}	& 34.44\%	& 40.51\%	& 42.95\%\\
    \bottomrule
  \end{tabular}
  \caption{Model performance on Chinese with various (B)WE initializations.}
  \label{tab:wordemb}
\end{table}

In this section we discuss the effect of the bilingual word embeddings.
We start by initializing the systems with random word embeddings (WEs), shown in Table~\ref{tab:wordemb}.
\texttt{ADAN} with random WEs outperforms the \texttt{DAN} and mSDA baselines using BWEs and matches the performance of the LR+MT baseline (Table~\ref{tab:adanexp}),
suggesting that \texttt{ADAN} successfully extracts features that could be used for cross-lingual classification tasks without \emph{any} bitext.
This impressive result vindicates the power of adversarial training to reduce the distance between two complex distributions without any direct supervision, which is also observed in other recent works for different tasks~\cite{zhang-EtAl:2017:Long5,lample2018word}.

With the introduction of BWEs (Column 2 and 3),
the performance of \texttt{ADAN} is further boosted.
Therefore, it seems the quality of the BWEs plays an important role in CLSC.
To investigate the impact of the specific choice of BWEs,
we also trained 100d BilBOWA BWEs~\cite{gouws2015bilbowa} using the UN parallel corpus for Chinese.
All systems achieve slightly lower performance compared to the pre-trained BWEs from \newcite{zou-EtAl:2013:EMNLP},
yet \texttt{ADAN} still outperforms other baseline methods (Table~\ref{tab:wordemb}),
demonstrating that \texttt{ADAN}'s effectiveness is relatively robust with respect to the choice of BWEs.
We conjecture that all systems show inferior results with BilBOWA, because it does not require word alignments during training as \newcite{zou-EtAl:2013:EMNLP} do.
By only training on a sentence-aligned corpus,
BilBOWA requires less resources and is much faster to train,
potentially at the expense of quality.

\subsubsection{Feature Extractor Architectures}\label{sec:f_arc_exp}
\begin{table}[bt]
  \centering
  \begin{tabular}{lcc}
    \toprule
    Model	&	Accuracy	 &	Run time \\
    \midrule
    \texttt{DAN}	& 42.95\% & 0.127 (s/iter)\\
    \texttt{CNN}	& 46.24\% & 0.554 (s/iter)\\
    \texttt{BiLSTM}	& 44.55\% & 1.292 (s/iter)\\
    \texttt{BiLSTM} + dot attn	& 46.41\% & 1.898 (s/iter) \\
    \bottomrule
  \end{tabular}
  \caption{Performance and speed for various feature extractor architectures on Chinese.}
  \label{tab:f_arch}
\end{table}

As mentioned in \secref{sec:modelarch}, the architecture of \adan{}'s feature extractor is not limited to a Deep Averaging Network (\texttt{DAN}), and one can choose different feature extractors to suit a particular task or dataset.
While an extensive study of alternative architectures is beyond the scope of this work, we in this section present a brief experiment illustrating that our adversarial framework works well with other $\cf$ architectures.
In particular, we consider two popular choices: i) a CNN~\cite{D14-1181} that has a 1d convolutional layer followed by a single fully-connected layer to extract a fixed-length vector;
and ii) a Bi-LSTM with two variants: one that takes the average of the hidden outputs of each token as the feature vector, and one with the dot attention mechanism~\cite{D15-1166} that learns a weighted linear combination of all hidden outputs.

As shown in Table~\ref{tab:f_arch}, \adan{}'s performance can be improved by adopting more sophisticated feature extractors, at the expense of slower running time.
This demonstrates that \adan{}'s language-adversarial training framework can be successfully used with other $\cf$ choices.

\subsubsection{\texttt{ADAN} Hyperparameter Stability}\label{sec:hyperparameter}
In this section,
we show that the training of \texttt{ADAN} is stable over a large set of hyperparameters,
and provides improved performance compared to the standard \adan{}-GRL.

\begin{figure}[t]
  \centering
  \includegraphics[width=0.48\textwidth]{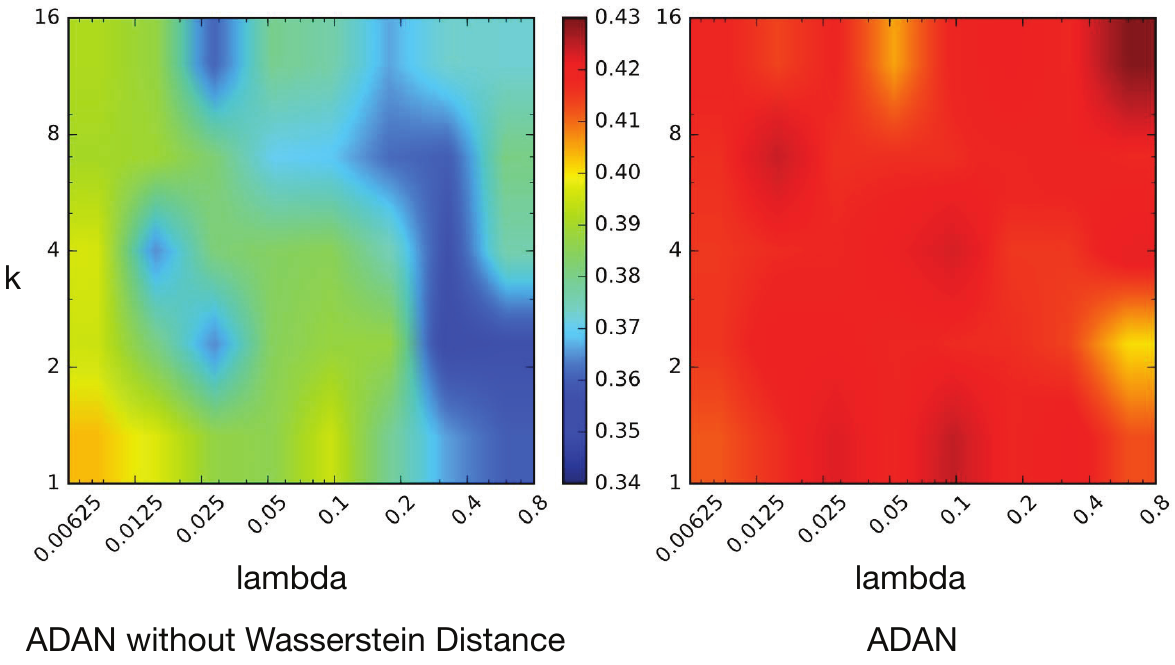}
  \caption{A grid search on $k$ and $lambda$ for \adan{} (right) and the \adan{}-GRL variant (left).
  Numbers indicate the accuracy on the Chinese development set.
  }
  \label{fig:gridsearch}
\end{figure}

To verify the superiority of \adan{},
we conduct a grid search over $k$ and $\lambda$,
which are the two hyperparameters shared by \adan{} and \adan{}-GRL.
We experiment with $k\in\{1,2,4,8,16\}$,
and $\lambda\in \{0.00625,0.0125,0.025,0.05,0.1,0.2,0.4,0.8\}$.
Figure~\ref{fig:gridsearch} reports the accuracy on the Chinese dev set for both \adan{} variants, and shows higher accuracy and greater stability over the \newcite{icml2015_ganin15} variant.
This suggests that \adan{} overcomes the well-known problem that adversarial training is sensitive to hyperparameter tuning.

\subsection{Implementation Details}
For all our experiments on both languages, the feature extractor $\cf$ has three fully-connected layers with ReLU non-linearities, while both $\cp$ and $\cq$ have two. 
All hidden layers contain 900 hidden units.
Batch Normalization~\cite{ioffe2015batch} is used in each hidden layer in $\cp$ and $\cq$.
$\cf$ does not use batch normalization.
$\cf$ and $\cp$ are optimized jointly using \texttt{Adam}~\cite{kingma2014adam} with a learning rate of $0.0005$.
$\cq$ is trained with another \texttt{Adam} optimizer with the same learning rate.
The weights of $\cq$ are clipped to $[-0.01, 0.01]$.
We train \texttt{ADAN} for 30 epochs and use early stopping to select the best model on the validation set.
\texttt{ADAN} is implemented in \texttt{PyTorch}~\cite{paszke2017automatic}.

\section{Related Work}
\noindent
\textbf{Cross-lingual Sentiment Classification} is motivated by the lack of high-quality labeled data in many non-English languages~\cite{10.1007/978-3-540-45175-4_13,mihalcea2007learning,banea2008multilingual,banea2010multilingual,soyer2014leveraging}.
For Chinese and Arabic in particular, there are several representative works~\cite{wan2008using,wan2009co,he2010exploring,lu2011joint,mohammadSK2015}.
Our work is comparable to these in objective but very different in method.
The work 
by Wan uses machine 
translation to directly convert English training data to Chinese; this is one of our baselines.
\newcite{lu2011joint} instead uses labeled data from both languages to improve the performance on both.
Other papers make direct use of a parallel corpus either to learn a bilingual document representation~\cite{P16-1133} or to conduct cross-lingual distillation~\cite{P17-1130}.
\newcite{P16-1133} require the translation of the entire English training set which is prohibitive for our setting, while \adan{} outperforms \newcite{P17-1130}'s approach in our experiments.

\noindent
\textbf{Domain Adaptation} tries to learn effective classifiers for which the training and test samples are from different underlying distributions~\cite{blitzer-dredze-pereira:2007:ACLMain,pan2011domain,glorot2011domain,ICML2012Chen_416,Liu:2015:IDS:2832415.2832427}.
This can be thought of as a generalization of cross-lingual text classification.
However, one main difference is that, when applied to text classification tasks,
most of these domain adaptation work assumes a common feature space such as a bag-of-words representation, which is not available in the cross-lingual setting.
See Section~\ref{sec:adanexp} for experiments on this.
In addition, most works in domain adaptation evaluate on adapting product reviews across domains (e.g.\ books to electronics), where the divergence in distribution is less significant than that between two languages.

\noindent
\textbf{Adversarial Networks} have enjoyed much success in computer vision~\cite{DBLP:conf/nips/GoodfellowPMXWOCB14,JMLR:v17:15-239}.
A series of work in image generation has used architectures similar to ours, by pitting a neural image generator against a discriminator that learns to classify real versus generated images \cite{DBLP:conf/nips/GoodfellowPMXWOCB14}.
More relevant to this work, adversarial architectures have produced the state-of-the-art in unsupervised domain adaptation for image object recognition: \newcite{JMLR:v17:15-239} train with many labeled source images and unlabeled target images, similar to our setup.
In addition, other recent work~\cite{arjovsky2017wasserstein,2017arXiv170400028G} proposes improved methods for training Generative Adversarial Nets.
\adan{} proposes \emph{language-adversarial training}, the first adversarial neural net for cross-lingual NLP~\cite{2016arXiv160601614C}.
As of the writing of this journal paper, there are several other recent works that adopt adversarial training for cross-lingual NLP tasks, such as cross-lingual text classification~\cite{P17-1130}, cross-lingual word embedding induction~\cite{zhang-EtAl:2017:Long5,lample2018word} and cross-lingual question similarity reranking~\cite{joty-EtAl:2017:CoNLL}.
\section{Conclusion and Future Work}
In this work, we presented \texttt{ADAN}, an adversarial deep averaging network for cross-lingual sentiment classification.
\texttt{ADAN} leverages the abundant labeled resources from English to help sentiment classification on other languages where little or no annotated data exist.
We validate \adan{}'s effectiveness by experiments on Chinese and Arabic sentiment classification,
where we have labeled English data and only \textit{unlabeled} data in the target language.
Experiments show that \texttt{ADAN} outperforms several baselines including domain adaptation models, a competitive MT baseline, and state-of-the-art cross-lingual text classification methods.
We further show that even without \emph{any} bilingual resources, \adan{} trained with randomly initialized embeddings can still achieve encouraging performance.
In addition, we show that in the presence of labeled data in the target language,
\texttt{ADAN} can naturally incorporate this additional supervision and yields even more competitive results.

For future work, we plan to apply our language-adversarial training framework to other NLP adaptation tasks
where explicit MLE training is not feasible due to the lack of direct supervision.
Our framework is not limited to sentiment classification or even to generic text classification:
It can be applied, for example, to phrase-level tagging tasks~\cite{irsoy-drnt} where labeled data might not exist for certain languages.
In another direction, we can look beyond a single \source{} and \target{} language and utilize our adversarial training framework for multi-lingual text classification.

\section*{Acknowledgments}
We thank the anonymous reviewers and members of Cornell NLP and ML groups for helpful comments.  This work was funded in part by a grant from the DARPA Deft Program.

\bibliography{adan}
\bibliographystyle{acl_natbib}
\end{document}